\documentclass[a4paper,conference]{IEEEtran}

\ifCLASSINFOpdf
\else
\fi

\usepackage{graphics} 
\usepackage{epsfig} 
\usepackage{mathptmx} 
\usepackage{times} 
\usepackage{amsmath} 
\usepackage{amssymb}  

\usepackage{subfigure}

\usepackage{algorithmicx}
\usepackage{algpseudocode}
\usepackage{booktabs}
\usepackage{multirow}
\usepackage{array}

\usepackage{amssymb}
\newcommand{\tabincell}[2]{\begin{tabular}{@{}#1@{}}#2\end{tabular}}
\usepackage{caption}
\usepackage{color}
\IEEEoverridecommandlockouts

\begin{document}
\title{Manual-Label Free 3D Detection via An Open-Source Simulator}

\author{\IEEEauthorblockN{Zhen Yang\textsuperscript{\rm 1,2}, Chi Zhang\textsuperscript{\rm 1,3}\IEEEauthorrefmark{1}\thanks{\textsuperscript{\rm *} corresponding author}\footnote{hhhhh}, Huiming Guo\textsuperscript{\rm 2}, Zhaoxiang Zhang\textsuperscript{\rm 1,3}}
\IEEEauthorblockA{\textit{\textsuperscript{\rm 1}Institute of Automation, Chinese Academy of Sciences, Beijing, China}\\
\textit{\textsuperscript{\rm 2}Beijing Aerospace Changfeng Co.Ltd., The $2^{nd}$ Institute of CASIC, Beijing, China}\\
\textit{\textsuperscript{\rm 3}Artificial Intelligence Research, Chinese Academy of Sciences, Jiaozhou, Qingdao, China}\\
yangzhen1324@163.com, chi.zhang@ia.ac.cn, ghm3651@vip.sina.com, zhaoxiang.zhang@ia.ac.cn}
}
\maketitle

\begin{abstract}
LiDAR based 3D object detectors typically need a large amount of detailed-labeled point cloud data for training, but these detailed labels are commonly expensive to acquire. In this paper, we propose a manual-label free 3D detection algorithm that leverages the CARLA simulator to generate a large amount of self-labeled training samples and introduces a novel Domain Adaptive VoxelNet (DA-VoxelNet) that can cross the distribution gap from the synthetic data to the real scenario. The self-labeled training samples are generated by a set of high quality 3D models embedded in a CARLA simulator and a proposed LiDAR-guided sampling algorithm. Then a DA-VoxelNet that integrates both a sample-level DA module and an anchor-level DA module is proposed to enable the detector trained by the synthetic data to adapt to real scenario. Experimental results show that the proposed unsupervised DA 3D detector on KITTI evaluation set can achieve 76.66\% and 56.64\% mAP on BEV mode and 3D mode respectively. The results reveal a promising perspective of training a LIDAR-based 3D detector without any hand-tagged label.
\end{abstract}
%
\IEEEpeerreviewmaketitle

\section{Introduction}
3D object detection can provide detailed spatial and semantic information, i.e., 3D position, orientation, occupied volume as well as category, and thus lures more and more attention in recent years. However, 3D object detection commonly needs a large amount of well-labeled data for training, and these detailed labels are commonly expensive to acquire.

\begin{figure}[t]
\centering
\includegraphics[width=\columnwidth]{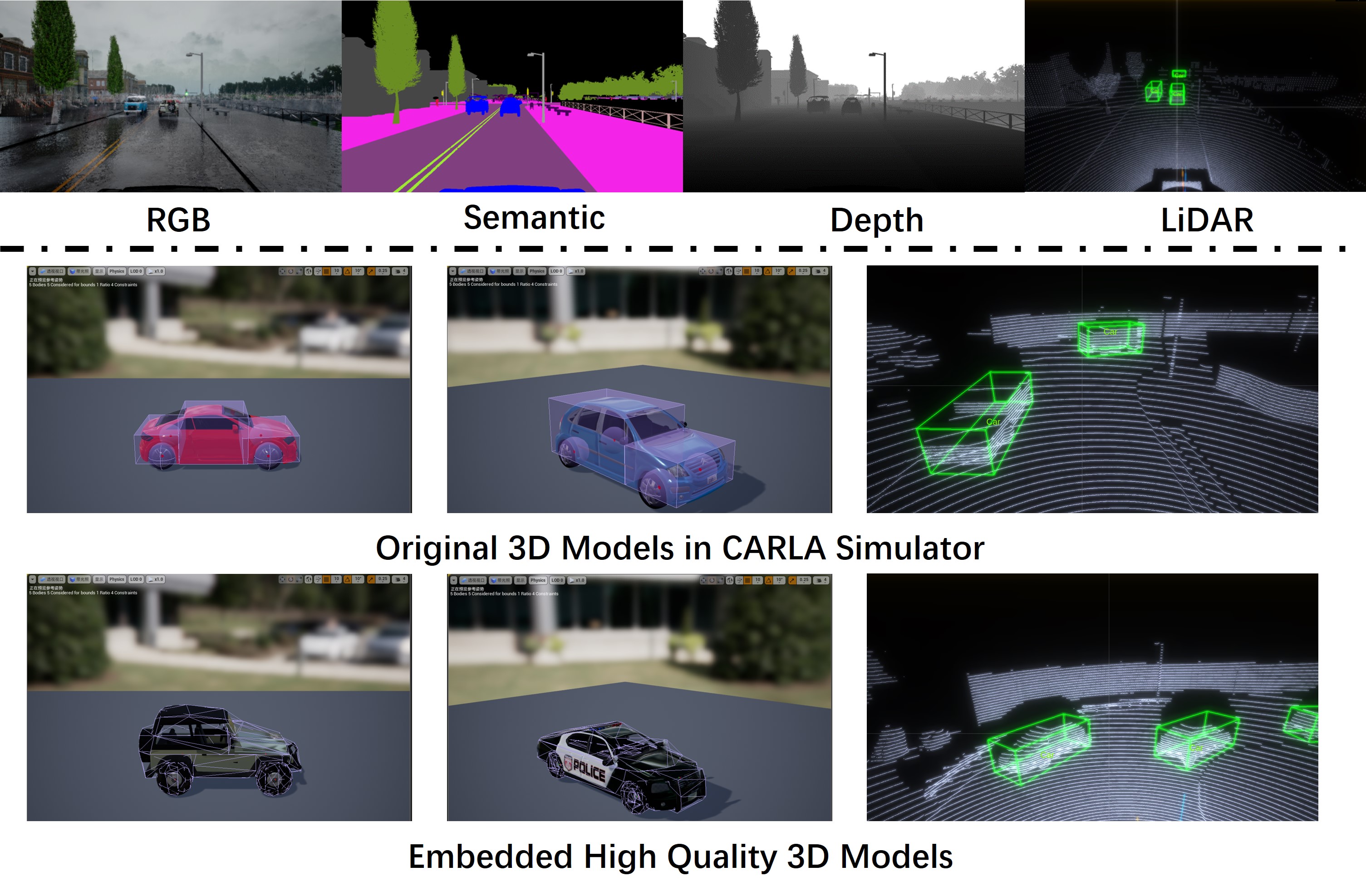}
\caption{Upper: Illustration of different sensors in CARLA. For the point cloud data, to facilitate the observation, we use a green bounding box to mark the object out. Lower: A comparison between the high quality 3D models embedded in the CARLA simulator and the original 3D models in CARLA simulator.}
\label{f3}
\end{figure}

\begin{figure*}[t]
\centering
\includegraphics[width=1.0\textwidth]{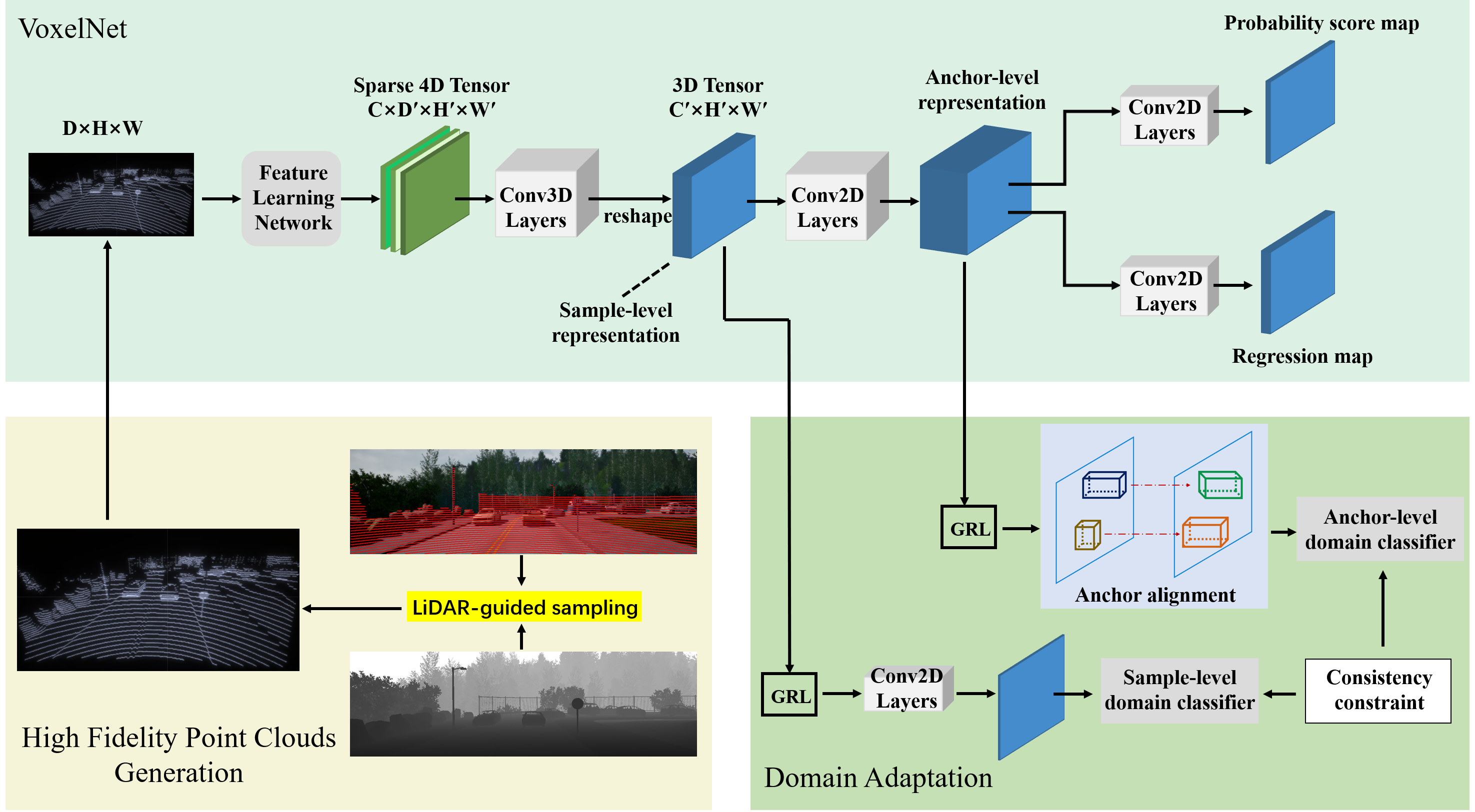}
\caption{An overview of our proposed method: we first propose a novel LiDAR-guided sampling algorithm to generate high fidelity point clouds. We then propose two novel domain adaptation components to cross the gap between the synthetic data and the real data, and further propose a consistency constraint to stabilize the training process.}
\label{f2}
\end{figure*}

Recently, the simulators are being increasingly used to remedy the shortage of labeled data. The simulators can generate self-labeled synthetic data with approximately no costs and have been used in multi-task learning \cite{ren2018cross}, semantic segmentation \cite{saleh2018effective}, multi-object tracking \cite{gaidon2016virtual}, etc.
In this work, we use the powerful CARLA simulator \cite{dosovitskiy2017carla} to generate self-labeled point cloud data. There are 4 types of sensor data offered by the stable version (0.8.2) of CARLA simulator (top of Fig.1). It should be mentioned that the physical model of the vehicle in CARLA simulator is simplified as several adjacent cuboids. As a result, the point cloud data obtained from these simplified models is very distorted. So, in order to get more realistic virtual point cloud data for the vehicle, we embed high quality 3D models into the CARLA simulator. The comparison between the embedded high quality 3D models and the CARLA simulator's original 3D models is illustrated in the bottom of Fig.1. Based on the sensors in the CARLA simulator, there are two methods to obtain the virtual point cloud data: 1) obtain the point cloud data from the rotating LiDAR implemented with ray-casting in CARLA, let's denote this method as CARLA-origin; 2) utilize the depth map and project the depth map back to the LiDAR's 3D coordinate system to get the pseudo point cloud data, let's denote this method as depth back-projection (depth-bp). As shown in Fig.3, the CARLA-origin method has the artefacts of the point cloud data generated by real LiDAR, but the shape of the vehicle is not realistic enough; the depth-bp method has a realistic shape of the vehicle, but loses the artefacts. In order to make the generated point cloud data have both the artefacts and the realistic vehicle appearance, we propose a novel sampling algorithm, LiDAR-guided sampling, to generate high fidelity point cloud samples from the depth map. By LiDAR-guided, we mean that the sampling process is guided by the spatial distribution of the point cloud data obtained from LiDAR.

However, the data generated by the simulator still has a visible difference comparing to the real data. And such discrepancies have been observed to cause significant performance drop \cite{gopalan2011domain}. The same is true for point cloud data, as shown in Table 1. To address the domain shift, we build an end-to-end trainable model based on the VoxelNet model \cite{zhou2018voxelnet}, referred to as Domain Adaptive VoxelNet (DA-VoxelNet). The DA-VoxelNet integrates two typical DA module to align features at sample-level and anchor-level respectively. The two DA modules are both constructed by a domain classifier and learned by an adversarial training manner. The adversarial training is implemented by introducing the gradient reversal layer (GRL) proposed in Ganin \emph{et al.} \cite{ganin2014unsupervised}. Given that the space dimension and the appearance of the object from different domains has a large variance, thus the domain classifier for the anchor-level adaptation is designed to look at the discrepancy of the bounding boxes from different domains. In addition, follow the consistency regularization proposed in Chen \emph{et al.} \cite{chen2018domain}, we further propose a consistency constraint to stabilize the domain classifier.

The main contributions of this work are summarized as follows: 1) We produce high quality 3D models and embed these models into CARLA simulator to get more realistic virtual point clouds. We then propose a novel sampling algorithm, LiDAR-guided sampling, to generate high fidelity point cloud samples. By utilizing the high fidelity point clouds to augment the training set, we can implement a promising 3D detector with exponentially reduced manual labeled data. 2) We propose two novel domain adaptation components to cross the gap between the synthetic data and the real data. We further impose a consistency constraint to stabilize the training process. Combine the both themes, the proposed DA-VoxelNet can get rid of the manual annotations thoroughly.

\section{Related Work}

\subsection{3D Object Detection in Point Cloud}
Recent 3D object detection methods start a new era of DNN-based 3D shape representation. Zhou \emph{et al.} \cite{zhou2018voxelnet} propose the voxel feature encoding (VFE) layer and use it to encode the point cloud.
In order to generate high quality proposals and fully exploit the detailed texture information provided by images, Ku \emph{et al.} \cite{ku2018joint} used a regional proposal network to simultaneously learn the features of the Birds Eye View (BEV) and images. For similar reasons, many methods \cite{ku2018joint,wang2019frustum,shin2018roarnet,qi2018frustum,chen2017multi} use both point cloud and RGB images as inputs to achieve performance gains. Although fusing both point clouds and RGB images can integrate more complimentary information, the LiDAR-only 3D detectors shows a comparable performance but significantly improved usability and robustness. So, in this work, we use the VoxelNet \cite{zhou2018voxelnet}, a LiDAR-only 3D detection method, as our baseline model.

Both the fused detectors and LiDAR-only detectors all need a great number of well-labeled training data that are commonly expensive to collect, especially for 3D data. Therefore, recently many research works begin to use simulators and simulated 3D LiDAR sensors for this task to solve this problem.

\subsection{Simulators}
Recently, many simulators are introduced to generate a large amount of self-labeled data for training DNN, like AirSim, Apollo, LGSVL and CARLA. AirSim \cite{shah2018airsim} is a simulator for drones, cars and more, built on Unreal Engine, and offers physically and visually realistic simulations. Apollo \cite{Apollo} is a flexible architecture which accelerates the development, testing, and deployment of Autonomous Vehicles. LGSVL \cite{LGSVL} is an HDRP Unity-based multi-robot simulator for autonomous vehicle developers. It is capable of rendering 128-beam LiDAR in real-time and can generate HD maps. CARLA \cite{dosovitskiy2017carla} is an open-source simulator for autonomous driving research and supports development, training, and validation of autonomous driving systems. It is powerful due to its ability to be controlled programmatically with an external client, which can control many aspects of the simulation process, from the weather to the vehicle route, and retrieve data from different sensors and send control commands to the player's vehicle.

\subsection{ Cross-Domain Object Detection}
We hope the detector can easily adapt to the real scenario after it was trained using synthetic data. But using an object detector to inference on a novel domain directly will cause a significant drop in performance \cite{wilber2017bam}.
Based on the deformable part-based model (DPM), Xu \emph{et al.} \cite{xu2014domain} use the adaptive structural SVM to enhance the adaptive ability of the DPM. Chen \emph{et al.} \cite{chen2018domain} use an adversarial training manner to learn a domain adaptive RPN for the Faster R-CNN model. Some recent works \cite{inoue2018cross,kim2019diversify,saleh2019domain} relied on the CycleGAN \cite{zhu2017unpaired} architecture try to overcome dataset bias by translating source domain images into the target style. In addition, with a large number of unlabeled videos from the target domain, Roychowdhury \emph{et al.} \cite{roychowdhury2019automatic} directly obtain labels on the target data by using an existing object detector and a tracker, the labels are then used for re-training the detector. Different from these works, we build a domain-adaptive and end-to-end trainable model for 3D object detection. To our best knowledge, the DA algorithm we proposed is the first of its kind.

\section{Proposed Method}
We implement a manual-label free 3D detection algorithm via an open-source simulator by two ways. On the one hand, we propose a novel LiDAR-guided sampling algorithm. It is capable to synthesize high-fidelity point cloud samples that can be used to train 3D detectors. On the other hand, we propose a novel unsupervised domain adaptation algorithm to align the domain shift of a 3D detector that learned from the virtual point clouds only.

\begin{figure}[t]
\centering
\subfigure[CARLA-origin]{\includegraphics[width=0.4\columnwidth]{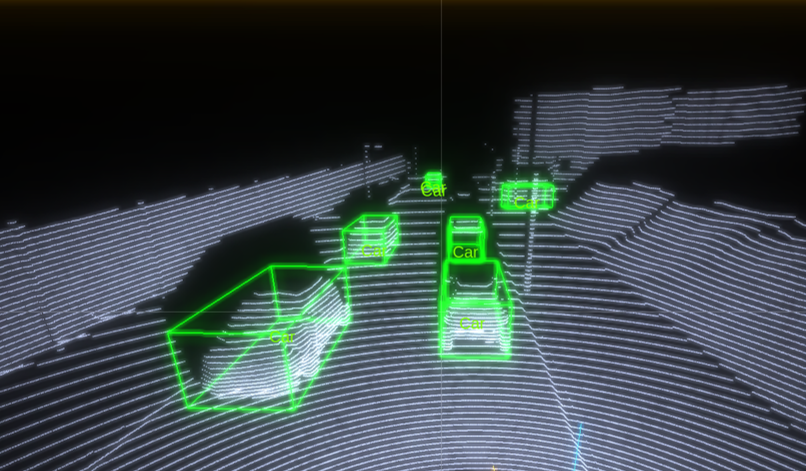}}
\quad
\subfigure[Depth-bp]{\includegraphics[width=0.4\columnwidth]{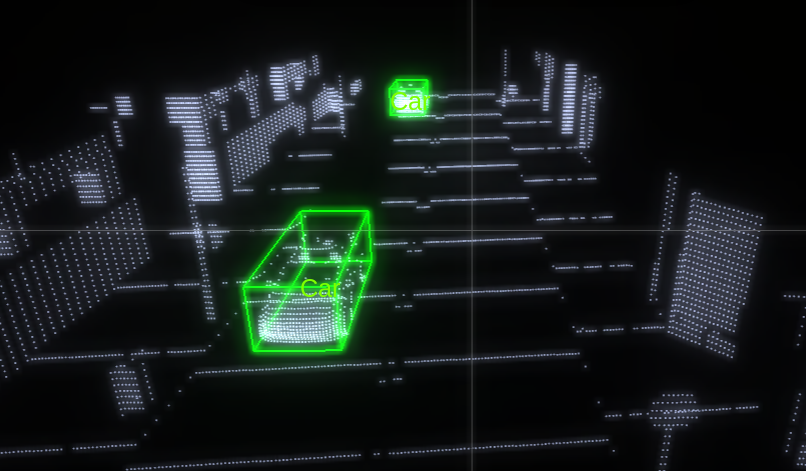}}
\quad
\subfigure[LiDAR-guided]{\includegraphics[width=0.4\columnwidth]{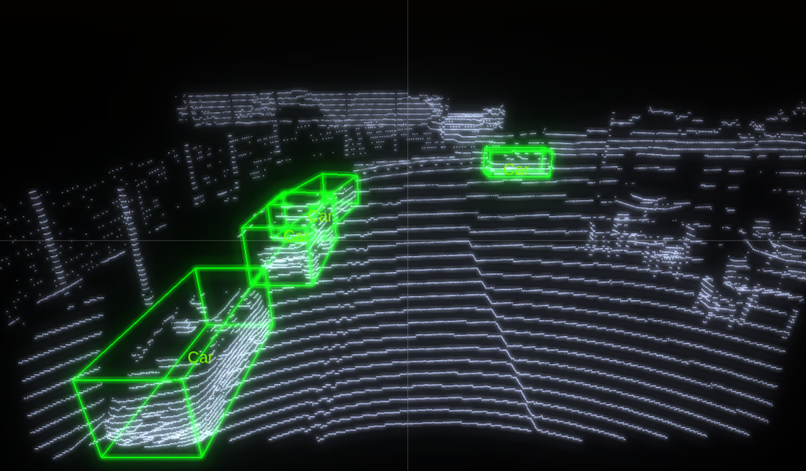}}
\quad
\subfigure[KITTI(real scene)]{\includegraphics[width=0.4\columnwidth]{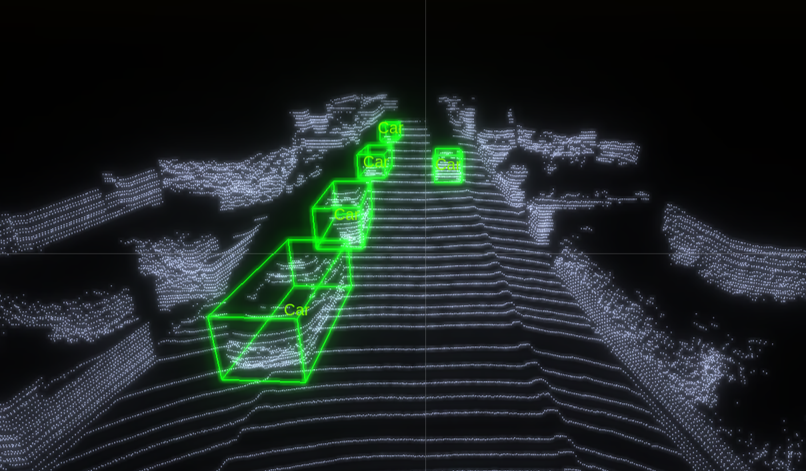}}
\caption{Comparison between the synthetic data generated with different method and the real data.}
\label{f5}
\end{figure}

\subsection{LiDAR-guided Sampling}
As shown in Fig.2, by LiDAR-guided, we mean that the sampling process is guided by the spatial distribution of the point cloud data obtained from LiDAR.
Specifically, we project the point clouds generated by LiDAR onto the 2D image plane, and we denote the set of point cloud from LiDAR as $\{ (u_i, v_i)\}^n_{i=1}\sim L$, where $(u,v)$ is the mapping coordinate of the point clouds on the 2D image plane. We also denote the pseudo point cloud from the depth map as $\{ (x_i, y_i)\}^m_{i=1}\sim D$, where $(x,y)$ is the mapping coordinate of the point cloud on the depth map. By building the identity mapping from 2D image plane to the depth map, we use the point clouds from LiDAR to guide the pseudo point clouds sampling. When $D_s$ denotes the sampled point cloud set from the pseudo point cloud set, we initialize it with $L$. We first connect adjacent points in $L$ along the LiDAR scan direction, and sample all points on the connected lines from corresponding positions in the pseudo point cloud to form the point set $D_{aug}$. For each point in $L$, the direction of the LiDAR scan is defined as the direction from this point to the closest point. We then compare the distance $d$ between two adjacent points in $D_{aug}$ along the direction of the LiDAR scan in turn. Concretely, we randomly pick a point which belong to the set $D_s\cup D_{aug}$ as the start point from the edge of the 2D image plane, and find a point closest to the start point as the next start point. If the two adjacent points all belong to $D_s$, there is no operations and we continue to next iteration. If one of the two adjacent points belongs to $D$, we compare the distance $d$ between the two points with the distance threshold parameter $d_{min}$. If $d\textless d_{min}$, we eliminate the next start point from $D_{aug}$ and reset the start point as next start point. In this way, we can control the density of the sampled point cloud data by adjusting the value of $d_{min}$. In this work, we set the $d_{min}$ of each point as the distance from this point to the closest point in $D_s$. After iterating each point in $D_{aug}$, we get a sampled subset of $D_{aug}$, denoted as $P_s$. Then, the high-fidelity point cloud set $P$ can be summarized as:

\begin{equation}
f(x, y, D)=
\begin{cases}
1,\quad (x, y)\in D\\
0,\quad (x, y)\not\in D
\end{cases}
\end{equation}

\begin{equation}
\delta (d)=
\begin{cases}
1,\quad d\geq d_{min}\\
0,\quad d\textless d_{min}
\end{cases}
\end{equation}

\begin{equation}
f(u,v, P_s)=\iint\limits_{HW}{f(x,y, D_{aug})\delta(\|(x,y), (u,v)\|_2)dxdy}
\end{equation}

\begin{equation}
P=D_s \cup P_s
\end{equation}

\noindent As shown in Fig.3, the virtual point cloud data generated by our proposed LiDAR-guided sampling algorithm is more realistic than other methods.


\begin{table*}[t]
\centering
\caption{The average precision (AP) of Car on the KITTI validation set and nuScenes validation set respectively. The VoxelNet is trained using the training set of LIDAR dataset (L), DEPTH dataset (D), CARLA dataset (C), KITTI (K) and nuScenes (nuS) as the source domain respectively. Among them, L, D and C are synthetic dataset generated by the CARLA simulator, K and nuS are collected from the real scene. {\color{red}\textbf{Red}} indicates the best and {\color{blue}\textbf{Blue}} the second best.}
\label{tab1}
\small
\begin{tabular}{c|c|ccc|c|ccc}
\cline{1-9}
 &Direction&Easy&Moderate&Hard&Direction&Easy&Moderate&Hard\\
\cline{1-9}
\multirow{4}*{BEV AP}&C$\rightarrow$K&61.09&53.13&49.30&C$\rightarrow$nuS&37.11&31.27&14.41\\
\cline{2-9}&D$\rightarrow$K&83.71&71.56&66.22&D$\rightarrow$nuS&50.82&40.75&19.40\\
\cline{2-9}&L$\rightarrow$K&{\color{blue}\textbf{87.71}}&{\color{blue}\textbf{74.89}}&{\color{blue}\textbf{68.01}}&L$\rightarrow$nuS&{\color{blue}\textbf{55.46}}&{\color{blue}\textbf{46.30}}&{\color{blue}\textbf{20.21}}\\
\cline{2-9}&K$\rightarrow$K&{\color{red}\textbf{89.97}}&{\color{red}\textbf{87.85}}&{\color{red}\textbf{86.84}}&nuS$\rightarrow$nuS&{\color{red}\textbf{74.82}}&{\color{red}\textbf{65.99}}&{\color{red}\textbf{31.67}}\\
\cline{1-9}
\multirow{4}*{3D AP}&C$\rightarrow$K&26.64&21.98&20.56&C$\rightarrow$nuS&2.08&1.79&1.30\\
\cline{2-9}&D$\rightarrow$K&68.09&52.84&46.00&D$\rightarrow$nuS&{\color{blue}\textbf{25.51}}&{\color{blue}\textbf{20.12}}&{\color{blue}\textbf{10.53}}\\
\cline{2-9}&L$\rightarrow$K&{\color{blue}\textbf{71.78}}&{\color{blue}\textbf{55.03}}&{\color{blue}\textbf{47.42}}&L$\rightarrow$nuS&23.78&18.32&9.75\\
\cline{2-9}&K$\rightarrow$K&{\color{red}\textbf{88.41}}&{\color{red}\textbf{78.37}}&{\color{red}\textbf{77.33}}&nuS$\rightarrow$nuS&{\color{red}\textbf{49.82}}&{\color{red}\textbf{42.24}}&{\color{red}\textbf{20.54}}\\
\cline{1-9}
\end{tabular}
\end{table*}

\subsection{Domain Adaptive VoxelNet}
We propose a novel unsupervised domain adaptation algorithm to further align the domain shift. We augment the VoxelNet base architecture with two domain adaptation components, which leads to our Domain Adaptive VoxelNet model (DA-VoxelNet). The architecture of our proposed DA-VoxelNet model is illustrated in Fig.2. We extract sample-level features and anchor-level features respectively, and perform sample-level adaptation and anchor-level adaptation accordingly. Following the consistency regularization proposed in Chen \emph{et al.} \cite{chen2018domain}, We further propose a consistency constraint to stabilize the training of domain classifiers.

\subsubsection{Sample-Level Adaptation}
In the VoxelNet model, the sample-level feature refers to the reshaped feature map which is the output of the 3D convolutional layers, as shown in Fig.2. To relieve the domain discrepancy caused by the ensemble difference of the point cloud sample such as density, regularity, etc., we utilize a domain classifier to align the target features with the source.

We have access to the labeled source point clouds $p^s$, as well as the unlabeled target point clouds $p^t$. The sample-level feature vector is extracted by $F_s$ and the anchor-level feature vector is extracted by $F_a$. The domain label is 0 for the source and 1 for the target. By denoting the domain classifier as $D_s$ and using the cross-entropy loss, the sample-level adaptation loss $L_{sample}$ can be summarized as:
\begin{equation}
L_{s_s}=-\frac{1}{n_s}\sum_{i=1}^{n_s}\log(1-D_s(F_s(p^s_i)))
\end{equation}
\begin{equation}
L_{s_t}=-\frac{1}{n_t}\sum_{i=1}^{n_t}\log(D_s(F_s(p^t_i)))
\end{equation}
\begin{equation}
L_{sample}=L_{s_s}+L_{s_t}
\end{equation}
where $n_s$ and $n_t$ denote the number of source and target examples respectively

In order to learn features that combine discriminative and domain-invariant representations, we should optimize the parameters of the base network to minimize the loss of the detection and to maximize the loss of the domain classifier. Meanwhile, to optimize the parameters of the domain classifier, we should also minimize the domain classification loss. In this way, the domain classifier should be optimized in an adversarial training manner. Such an adversarial training manner can be implemented by introducing the gradient reversal layer (GRL) \cite{ganin2014unsupervised,ganin2016domain}. For the base network, the GRL change the sign of the gradient by multiplying it with a certain negative constant $r$ during the backpropagation. For the domain classifier, the GRL performs identity mapping and the normal gradient descent is applied.

\subsubsection{Anchor-Level Adaptation}
The architecture of the anchor-level domain classifier, $D_a$, is designed to concentrate on the anchor-level features and perform anchor-level alignment. The anchor-level feature refers to the feature vectors which fed into the parallel convolutional layers to generate the final score map and the regression map, as shown in Fig.2. Because the bounding box of the 3D detections is based on the anchors, and the space dimension and the appearance of the object from different domains has a large variance. Therefore, it is crucial to align the anchors from the source domain to the target domain. To this end, we build an anchor-level domain classifier to obtain domain-invariant anchor-level features. Matching the anchor-level features helps to reduce the domain bias caused by instance diversity such as object appearance, space dimension, orientation, etc. Similar to the sample-level adaptation, using the cross-entropy loss, the anchor-level adaptation loss $L_{anchor}$ can be summarized as:
\begin{equation}
L_{ac_s}=-\frac{1}{n_s}\sum_{i=1}^{n_s}\log(1-D_a(F_a(p^s_i)))
\end{equation}
\begin{equation}
L_{ac_t}=-\frac{1}{n_t}\sum_{i=1}^{n_t}\log(D_a(F_a(p^t_i)))
\end{equation}
\begin{equation}
L_{anchor}=L_{ac_s}+L_{ac_t}
\end{equation}
We also use the GRL to make the domain classifier $D_a$ optimized in an adversarial training manner.

\subsubsection{Consistency Constraint}
To reinforce the domain classifiers and stabilize the training process, we further impose a consistency constraint between the domain classifiers on different levels. Let us assume the $F_s$ outputs a feature whose width and height is $W_s$ and $H_s$ and $F_a$ outputs a feature whose width and height is $W_a$ and $H_a$, we denote the loss of the consistency constraint as $L_{con}$ as follows,
\begin{equation}
M_s(n,p_i)=\frac{1}{n}\sum_{i=1}^{n}\sum_{h=1}^{H_s}\sum_{w=1}^{W_s}D_s(F_s(p_i))_{(w,h)}
\end{equation}
\begin{equation}
M_a(n,p_i)=\frac{1}{n}\sum_{i=1}^{n}\sum_{h=1}^{H_a}\sum_{w=1}^{W_a}D_a(F_a(p_i))_{(w,h)}
\end{equation}
\begin{equation}
L_{con_f}(n,p_i)=\|M_s(n,p_i) - M_a(n,p_i)\|_2
\end{equation}
\begin{equation}
L_{con}=L_{con_f}(n_s,p_i^s)+L_{con_f}(n_t,p_i^t)
\end{equation}
where $D_s(F_s(p_i))_{(w,h)}$ and $D_a(F_a(p_i))_{(w,h)}$ denotes the output of the domain classifier in each location.

\subsubsection{Overall Objective}
We denote the loss of detection modules as $L_{det}$, which contains the loss for classification and localization. The final training loss of the proposed network can be written as:
\begin{equation}
L=L_{det}+\lambda(L_{sample}+L_{anchor}+L_{con})
\end{equation}
where $\lambda$ controls the balance between the detection loss and the domain adaptation loss.

\section{Experiments}

\subsection{Datasets}

\subsubsection{Real Dataset}
KIITI \cite{geiger2012we} is a challenging benchmark for many tasks like optical flow, SLAM and 3D object detection et al. KITTI offers 7581 samples and more than 200k annotations for 3D objects detection. In this paper, we separate the 7581 samples into two non-overlap sets, the training set has 3721 samples and the validation set has 3769 samples.

nuScenes \cite{caesar2019nuscenes} offers a novel challenging benchmark for 3D object detection. nuScenes comprises 1000 scenes and fully annotated with 3D bounding boxes for 23 classes and 8 attributes. In this work, we concentrate on the detection for the Car. For convenience comparison with KITTI, we utilize the $num\_lidar\_pts$ provided by nuScenes to divide the data into easy, moderate and hard samples.

\subsubsection{Virtual Dataset}
We utilize the 0.8.2 version of CARLA simulator to generate the virtual point clouds. It is hard to acquire the precise bounding box for the person model because the person model in CARLA simulator has a physical defect. So, we only have one category of the object from the virtual point clouds generated by the CARLA simulator, which is vehicles.
We use the LiDAR-guided sampling algorithm to generate the virtual dataset from CARLA simulator, called LIDAR dataset. The CARLA dataset and DEPTH dataset correspond to the CARLA-origin method and the depth-bp method respectively. In the process of generating LIDAR dataset, CARLA dataset and DEPTH dataset, we set the same parameters in the simulator.
There are two default maps in CARLA simulator, named Town01 and Town02. For Town01, we set the number of car and walker as 100 and 50. Given that the size of the Town02 is small, we halved the number of the car and walker. For now, there are 17 unique 3D car models in CARLA simulator and we simulated three lighting conditions and two weathers, including morning, noon and afternoon, sunny and rainy days. There is an average of 3.41 cars in LIDAR dataset per scan, an average of 97989.86 points in LIDAR dataset per scan.
We offer 7000 samples and pick 6000 samples randomly to compose the training set, the rest is used for validation.

\subsection{Metrics}
In the following experiments, BEV AP and 3D AP are short for the average precision (in \%) of the bird’s eye view detection and 3D detection respectively. We follow the official KITTI evaluation protocol, where the IoU threshold is 0.7 for class Car.

\begin{table*}[t]
\centering
\caption{Quantitative analysis of finetune result from synthetic data to real data. $Percentage$ denotes the number of sampled data as a percentage of the target training set. If $Finetune$, we use the synthetic data to train the model and use the sampled data to finetune the model, else we use the sampled data to train the model directly.}
\label{tab3}
\small
\begin{tabular}{c|c|c|c|ccc|c|ccc}
\cline{1-11}
&Percentage&Finetune&Direction&Easy&Moderate&Hard&Direction&Easy&Moderate&Hard\\
\cline{1-11}
\multirow{5}{*}{BEV AP}&1\%&$\times$&K$\rightarrow$K&29.12&23.61&18.08&nuS$\rightarrow$nuS&34.52&27.54&13.96\\
\cline{2-11}&1\%&$\checkmark$&L$\rightarrow$K&89.49&79.22&77.95&L$\rightarrow$nuS&70.88&61.70&29.25\\
\cline{2-11}&5\%&$\checkmark$&L$\rightarrow$K&89.75&85.68&79.33&L$\rightarrow$nuS&72.80&63.78&29.79\\
\cline{2-11}&10\%&$\checkmark$&L$\rightarrow$K&\textbf{90.24}&86.71&86.31&L$\rightarrow$nuS&73.94&64.67&30.19\\
\cline{2-11}&100\%&$\times$&K$\rightarrow$K&89.97&\textbf{87.85}&\textbf{86.84}&nuS$\rightarrow$nuS&\textbf{74.82}&\textbf{65.99}&\textbf{31.67}\\
\cline{1-11}
\multirow{5}{*}{3D AP}&1\%&$\times$&K$\rightarrow$K&10.72&10.65&7.57&nuS$\rightarrow$nuS&7.21&5.03&2.49\\
\cline{2-11}&1\%&$\checkmark$&L$\rightarrow$K&83.82&72.25&66.00&L$\rightarrow$nuS&39.04&29.98&15.35\\
\cline{2-11}&5\%&$\checkmark$&L$\rightarrow$K&87.02&75.55&68.45&L$\rightarrow$nuS&46.26&38.15&18.53\\
\cline{2-11}&10\%&$\checkmark$&L$\rightarrow$K&87.51&76.40&74.45&L$\rightarrow$nuS&49.44&41.27&19.60\\
\cline{2-11}&100\%&$\times$&K$\rightarrow$K&\textbf{88.41}&\textbf{78.37}&\textbf{77.33}&nuS$\rightarrow$nuS&\textbf{49.82}&\textbf{42.24}&\textbf{20.54}\\
\cline{1-11}
\end{tabular}
\end{table*}

\subsection{LiDAR-guided Sampling}
To generate the high fidelity point cloud samples, we propose the LiDAR-guided sampling algorithm. The comparision between our proposed method and other methods is recorded in Tabel 1. Although the 3D detector trained using the DEPTH dataset performes better than using LIDAR dataset under the 3D AP evaluation criterion on the nuScenes dataset, the gap is not large. And the 3D detector trained using the LIDAR dataset outperforms the 3D detector trained using the DEPTH dataset by a large margin in other items. The results show clearly that the virtual point cloud data obtained by using our proposed LiDAR-guided sampling method has better generalization ability on real scenes.

Not like the image data that has a large variance in image style, image scale and illumination, etc., the point clouds have less variance. For this reason, we choose to use the high fidelity point clouds generated by CARLA simulator to train the VoxelNet directly, and expect to get a promising performance on the real data. Then, we use the KITTI training set and nuScenes training set as the training data respectively to compare the performance difference. The results for learning from synthetic data and the real data are recorded in Table 1. When we change the training set from real data to synthetic data, the performance of the detector drops to a large extent, from 77.33\% to 47.42\% at the most. The phenomenon indicates that even for point cloud data, there still exist a large domain gap from virtual data to real data.

In order to deal with the problem of domain shift between a source domain and a target domain, a simplest solution is to use the source domain data to train the model and use the labeled data from the target domain to finetune the model.  In this way, the model is enabled to learn discriminative representations for the target domain directly. We first use the virtual point clouds to pre-train the model. During training, we use Adam \cite{kingma2014adam} with learning rate 0.003 and a one cycle learning rate schedule \cite{smith2017cyclical} is adopted, a momentum of 0.9 and a weight decay of 0.001 is used in our experiments, the batch size is 8 and the training process iterates 50 epochs. Then we finetune the network on the target data for another 50 epoch iterations, with a learning rate of 0.0003.

As shown in Table 2, with the high-fidelity point cloud data, we only need 10\% labeled target data to get the performance that is comparable to the detector trained on the target domain. Especially for the evaluation in bird’s eye view on the KITTI validation set, we only pick 37 (10\%) samples from the KITTI training set randomly to finetune the detector, but outperforms the detector trained with the KITTI training set in $easy$ level. The results show clearly that the high fidelity point clouds we proposed is effective for reducing the amount of the manual annotations.

\begin{table}[t]
\centering
\caption{Results on adaptation from LIDAR to KITTI Dataset. Average precision (AP) of Car is evaluated on the KITTI validation set. $bs$ is short for batch size.}
\label{tab4}
\small
\begin{tabular}{c|c|c|ccc}
\cline{1-6}
 &bs&method&Easy&Moderate&Hard\\
\cline{1-6}
\multirow{4}{*}{BEV AP}&
\multirow{2}{*}{2}&VoxelNet&79.27&66.72&63.33\\
\cline{3-6}&&DA-VoxelNet&81.19&71.27&65.18\\
\cline{2-6}
&\multirow{2}{*}{8}&VoxelNet&87.71&74.89&68.01\\
\cline{3-6}&&DA-VoxelNet&\textbf{88.40}&\textbf{76.66}&\textbf{74.07}\\
\cline{1-6}
\multirow{4}{*}{3D AP}&
\multirow{2}{*}{2}&VoxelNet&57.04&43.02&40.62\\
\cline{3-6}&&DA-VoxelNet&65.18&51.61&45.10\\
\cline{2-6}
&\multirow{2}{*}{8}&VoxelNet&71.78&55.03&47.42\\
\cline{3-6}&&DA-VoxelNet&\textbf{73.77}&\textbf{56.64}&\textbf{52.29}\\
\cline{1-6}
\end{tabular}
\end{table}

\begin{table}[t]
\centering
\caption{Ablation study: Quantitative results on the KITTI validation set for $Moderate$ level, reported as mean and standard deviation over 3 rounds of training with batch size 2. Models are trained on the LIDAR training set. $an$ is short for anchor-level adaptation, $sa$ for sample-level adaptation and $cons$ is short for our consistency constraint.}
\label{tab5}
\small
\begin{tabular}{c|ccc|cc}
\cline{1-6}
method&sa&an&cons&\tabincell{c}{BEV AP\\ (mean$\pm$std)}&\tabincell{c}{3D AP\\ (mean$\pm$std)}\\
\cline{1-6}
VoxelNet&&&&66.44$\pm$0.43&43.42$\pm$0.62\\
\cline{1-6}
\multirow{4}{*}{DA-VoxelNet}&$\checkmark$&&&69.49$\pm$1.70&48.12$\pm$1.16\\
\cline{2-6}&&$\checkmark$&&70.50$\pm$0.20&50.30$\pm$0.42\\
\cline{2-6}&$\checkmark$&$\checkmark$&&70.92$\pm$0.17&50.15$\pm$0.68\\
\cline{2-6}&$\checkmark$&$\checkmark$&$\checkmark$&\textbf{71.15}$\pm$\textbf{0.33}&\textbf{50.57}$\pm$\textbf{1.61}\\
\cline{1-6}
\end{tabular}
\end{table}

\subsection{Domain Adaptive VoxelNet}
The finetune method is straightforward and effective, but requires a certain amount of labeled data from the target domain, the condition is unable to meet at times. In order to skip the manual labeling process, we propose the Domain Adaptive VoxelNet (DA-VoxelNet), an unsupervised DA learning method.
In this section, we use Adam with learning rate 0.003 and a one cycle learning rate schedule is adopted, a momentum of 0.9 and a weight decay of 0.001 is used in our experiments. In addition, we replace the batch normalization \cite{ioffe2015batch} with the group normalization \cite{wu2018group} when the batch size is reduced from 8 to 2.
During the training iteration, each batch is composed of 2 pair samples, one pair from the source domain and another pair from the target domain. We reported the performance trained after 30K iterations with batch size 2 and the performance trained after 50 epochs with batch size 8.

The results of the different methods are summarized in Table 3. Among them, our proposed method outperforms the VoxelNet model by a large margin. Specifically, when batch size is 2, DA-VoxelNet outperforms the VoxelNet by 1.92\%, 4.55\% and 1.85\% in $easy$, $moderate$ and $hard$ levels for BEV AP, 8.14\%, and 8.59\% and 4.48\% in $easy$, $moderate$ and $hard$ levels for 3D AP. When batch size is 8, DA-VoxelNet outperforms the VoxelNet by 0.69\%, 1.77\% and 6.06\% in $easy$, $moderate$ and $hard$ levels for BEV AP, 1.99\%, and 1.61\% and 4.87\% in $easy$, $moderate$ and $hard$ levels for 3D AP. We can see that when the batch size is small, the performance improvement benefits from the DA module is more obvious (up to 8.59\% when the batch is 2 and up to 4.87\% when the batch is 8). In addition, as shown in Table 4, we supplement the ablation study to verify the effectiveness of the various DA components we proposed. Given that the effect of the DA module is more distinct when batch size is small, we set the batch size to 2 when conducting the ablation study. Concretely, treat VoxelNet as a baseline model, the baseline model achieves a BEV AP of 66.44\% and a 3D AP of 43.42\% in $moderate$ level. With only sample-level adaptation, the performance is boosted to 69.49\% and 48.12\%. With only anchor-level adaptation, the performance is boosted to 70.50\% and 50.30\%. Combining the two levels adaptation together yields a BEV AP of 71.09\% (70.92+0.17) and a 3D AP of 50.83\% (50.15+0.68) at most. Finally, with the consistency constraint, the DA-VoxelNet achieves a BEV AP of 71.48\% and a 3D AP of 52.18\% at most. Among the DA components we proposed, the domain classifier of the anchor-level adaptation module is designed to focus on the discrepancy of the bounding boxes from different domains. As shown in Fig.4, with the anchor-level adaptation, we can reduce bounding-box bias caused by the diversity such as object space dimension and object orientation significantly.

\begin{figure}[t]
\centering
\subfigure[without anchor-level adaptation]{\includegraphics[width=0.9\columnwidth]{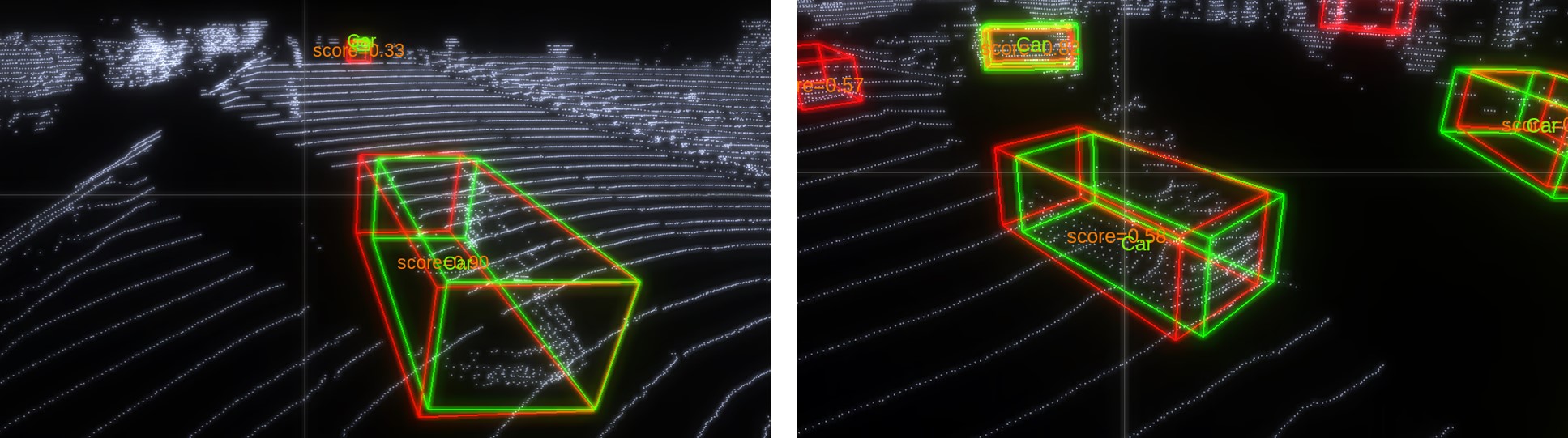}}
\quad
\subfigure[with anchor-level adaptation]{\includegraphics[width=0.9\columnwidth]{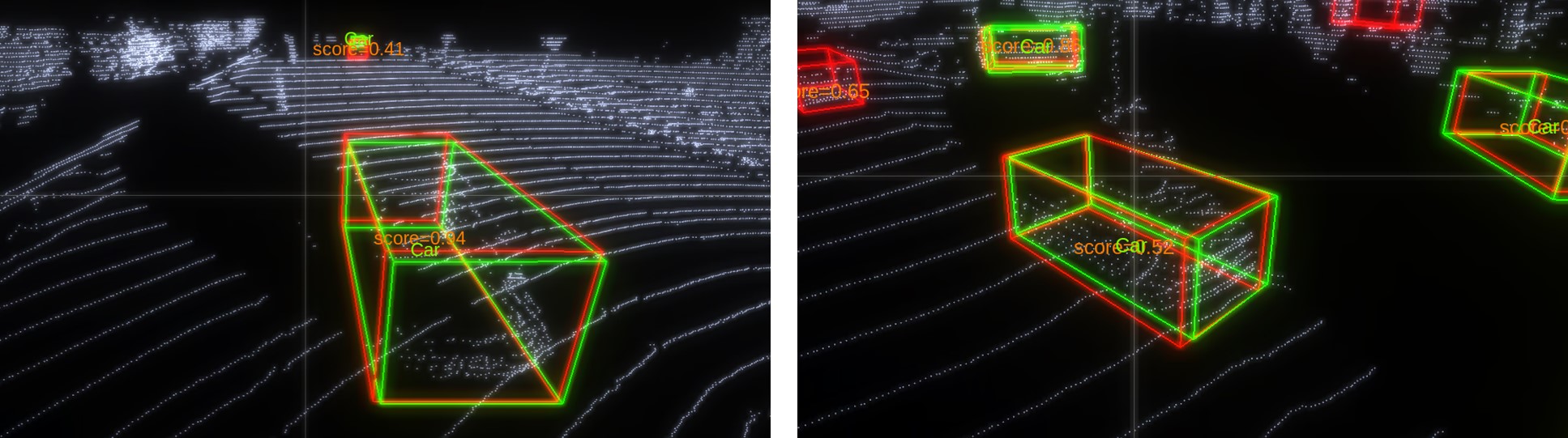}}
\quad
\caption{Visualization of inference results on KITTI validation set. The green bounding box represent ground truth and the red is the inference results.}
\label{f7}
\end{figure}

\section{Conclusion}
In this paper, we propose a manual-label free 3D detection algorithm. We embed high quality 3D models into the CARLA simulator and propose a novel LiDAR-guided sampling algorithm to generate high fidelity point cloud samples. By utilizing the high fidelity point cloud samples, We can exponentially reduce the amount of manual labeled data required to train a 3D object detector. We then propose a DA-VoxelNet that integrates both a sample-level DA module and an anchor-level DA module to enable the detector trained by the synthetic data to adapt to real scenario. Experimental results show that the DA-VoxelNet gain a large performance improvement compared to the VoxelNet(from $\sim$ 4\% improvement in bird's eye view to $\sim$ 8\% improvement on 3D detection), which reveals a promising perspective of training a LIDAR-based 3D detector without any hand-tagged label.

\section*{Acknowledgement}
This work was supported in part by the Major Project for New Generation of AI under Grant No. 2018AAA0100400, the National Natural Science Foundation
of China (No. 61836014, No. 61761146004, No. 61773375, No. 61602481).

\bibliographystyle{IEEEtran}
\bibliography{mybibfile}

\end{document}